\documentclass{article}
\usepackage{spconf,amsmath,graphicx}
\usepackage{multirow}
\usepackage{newtxmath}
\usepackage{booktabs} 
\usepackage{url} 

\title{Recouple Event Field via Probabilistic Bias for Event Extraction}
\name{
\begin{tabular}{c}
\it Xingyu Bai\textsuperscript{1*}, 
        Taiqiang Wu\textsuperscript{1*}\thanks{* Equal contribution.},
        Han Guo\textsuperscript{2},
        Zhe Zhao\textsuperscript{2},
        Xuefeng Yang\textsuperscript{2},\\
\it Jiayi Li\textsuperscript{1},
        Weijie Liu\textsuperscript{2},
        Qi Ju\textsuperscript{2},
        Weigang Guo\textsuperscript{2},
        Yujiu Yang\textsuperscript{1} $^\dagger$ \thanks{ $^\dagger$ Corresponding author. This work was partly supported by the National Key Research and Development Program of China (No. 2020YFB1708200) and the Shenzhen Key Laboratory of Marine IntelliSense and Computation under Contract ZDSYS20200811142605016.}
\end{tabular}
}
        
\address{\textsuperscript{1}Tsinghua Shenzhen International Graduate School, Tsinghua University, China\\
\textsuperscript{2}Tencent, China}
    
\usepackage{bibentry}
\begin{document}
%
\maketitle
\begin{abstract}
Event Extraction (EE), aiming to identify and classify event triggers and arguments from event mentions, has benefited from pre-trained language models (PLMs). However, existing PLM-based methods ignore the information of trigger/argument fields, which is crucial for understanding event schemas. To this end, we propose a Probabilistic reCoupling model enhanced Event extraction framework (ProCE). Specifically, we first model the syntactic-related event fields as probabilistic biases, to clarify the event fields from ambiguous entanglement. Furthermore, considering multiple occurrences of the same triggers/arguments in EE, we explore probabilistic interaction strategies among multiple fields of the same triggers/arguments, to recouple the corresponding clarified distributions and capture more latent information fields. Experiments on EE datasets demonstrate the effectiveness and generalization of our proposed approach.
\end{abstract}

\section{Introduction}
\label{sec:introduction}



Event extraction is an essential information extraction (IE) task, aiming to extract event structures from unstructured event mentions. It consists of event detection (ED) and event argument extraction (EAE). For example, in the mention "CNN's Kelly reports on Netanya's attack.", the ED model should identify the event trigger "attack" and classify event type "Conflict", the EAE model should identify the event argument "Netanya" and classify argument role "Place".


\begin{figure}[htb!]
\begin{center}
\includegraphics[width=0.95\linewidth]{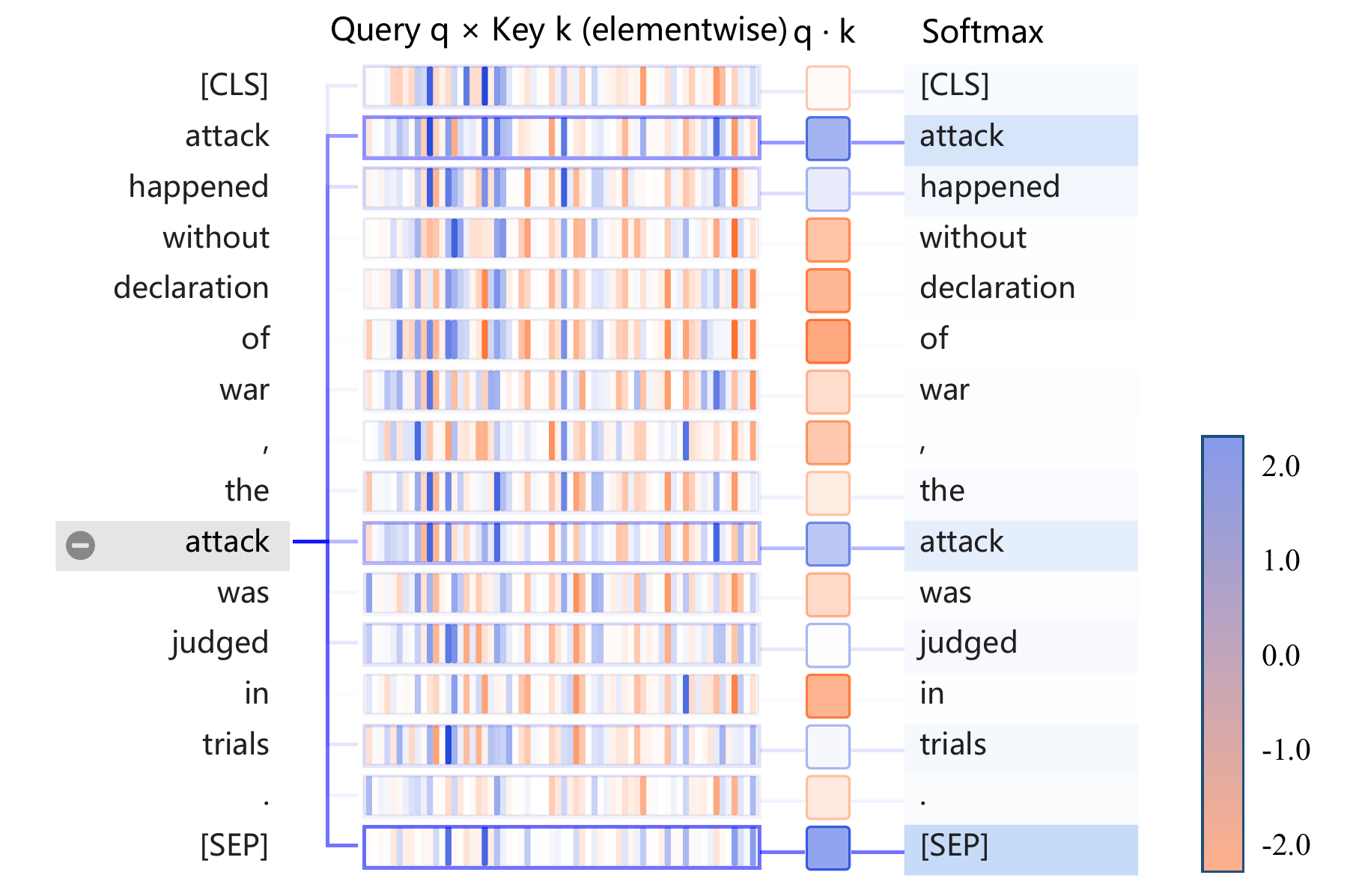}
\caption{Attention score in original self-attention entanglement, connecting line colors are weighted by attention values.} \label{attack}
\end{center}
\vspace{-2em}
\end{figure}

Existing EE methods mainly focus on feature engineering. Inspired by the significant performance of PLMs, some prior work \cite{wang2019adversarial,wadden2019entity} utilize general PLMs, such as BERT \cite{devlin2019bert}, to construct global dependencies among context words by self-attention. However, PLMs suffer from handling the entanglement of triggers/arguments. As shown in Figure~\ref{attack}, in the sentence "Attack happened without declaration of war, the attack was judged in trials.", the first trigger "attack" is more important to the second trigger "attack" than other words when computing self-attention, the information of trigger fields, which is defined as the core auxiliary information scope of triggers, is not highlighted and strengthened enough from the original self-attention entanglement.


To avoid the problem of auxiliary information insufficiency, dependency tree based Graph Convolution Network (GCN) \cite{nguyen2018graph,liu2018jointly,DBLP:conf/wsdm/WuBG0LY23} was adopted to capture syntactic relations between triggers and related words. However, this method still has some problems: (1) GCN focuses on the nearest syntactic neighbors ~\cite{nguyen2018graph,liu2018jointly}, over-smoothing \cite{zhou2020graph} in deep layers limits the message passing; (2) GCN has not modeled the event fields and entanglement among the identical event fields.



Another localness-enhanced method, which models central word regions as Gaussian priors \cite{yang2018modeling,guo2019gaussian}, could expand the neighbor scopes without changing the model structure.


\begin{figure*}[tb!]
\begin{center}
\includegraphics[width=0.9\textwidth]{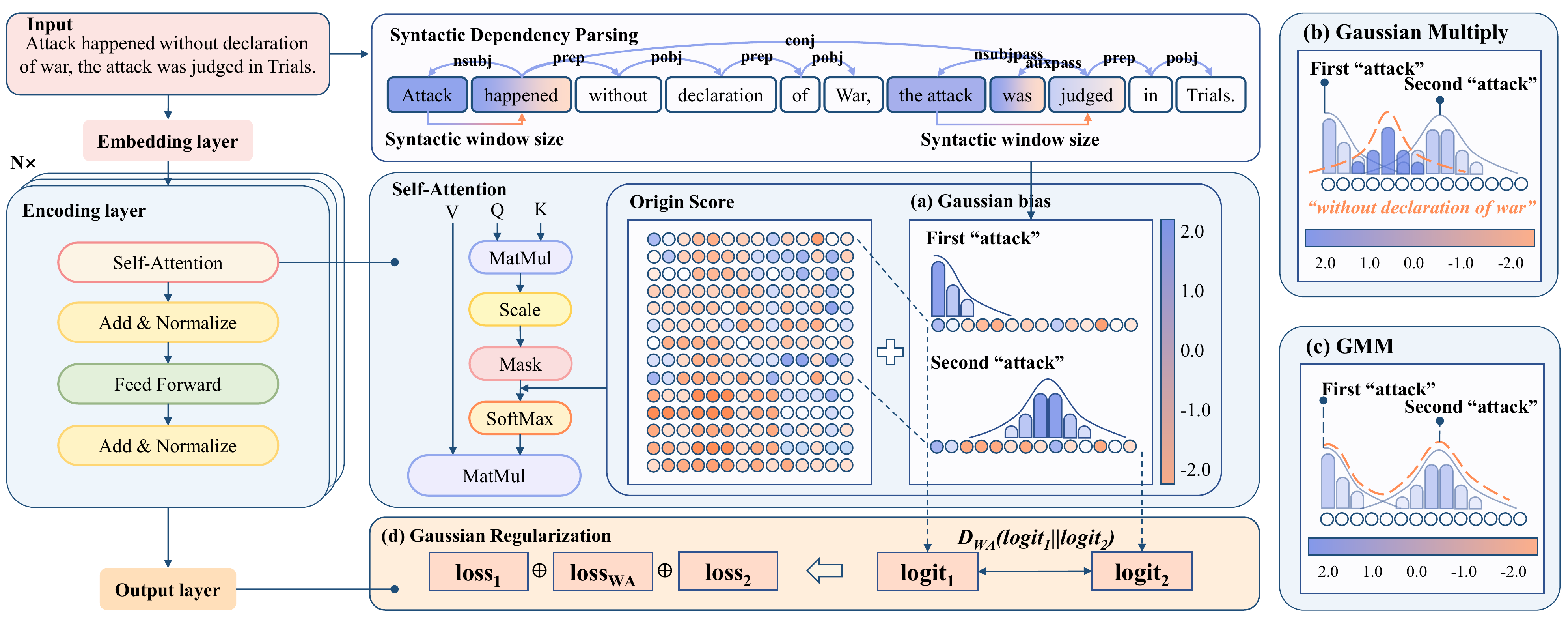}
\caption{Illustration of the proposed probabilistic-based event field modeling approach.} \label{model}
\end{center}
\vspace{-2em}
\end{figure*}

In this paper, we propose a novel probabilistic recoupling model enhanced EE framework. Specifically, we model and clarify the event fields using probabilistic priors over the corresponding window (i.e., the deviation of syntactic dependency distribution) to the central word (i.e., the position of trigger/argument), and further decouple the original ambiguous self-attention entanglement for the first time.


According to our statistics on the ACE-2005 dataset, 13.18\% (1959/14862) of mentions have the same triggers and arguments occurring multiple times. The same triggers/arguments in the same mentions are identical, modeling the entanglement among multiple occurrences of the same triggers/arguments is beneficial. We explore several probabilistic interaction methods among identical trigger/argument fields to recouple the clarified fields and capture more latent knowledge fields. In addition, we adopt a novel distribution metric loss and regularize the output distributions of the identical triggers/argument fields to be consistent by minimizing the Wasserstein (WA) divergence among the outputs. Experiments on several datasets indicate that ProCE achieves significant performances on overall and few-sample settings.

\section{Methodology}


The overall ProCE framework consists of three enhancement strategies: (1) Event Field Clarification Modeling, (2) Event Field Recoupling, (3) Event Field Regularization Learning.


\noindent{\textbf{\underline{Preprocessing.}}} To build syntactic-aware event fields, we use the syntactic dependency parser spaCy\footnote{\url{https://spacy.io/api/dependencyparser}} to parse event mentions into syntactic dependency tree structures, which consist of tokens as nodes and syntactic dependencies as edges.


\noindent{\textbf{\underline{Event Field Clarification Modeling.}}} Self-attention computes attention scores as Figure~\ref{attack}. Specifically, given an event mention $s=\{w_1,...,w_n\}$ contains $n$ tokens, the hidden state $H$ is calculated by the transformed queries $Q \in \mathbb{R}^{n \times d}$, keys $K \in \mathbb{R}^{n \times d}$, and values $V \in \mathbb{R}^{n \times d}$ as follows:
\vspace{-1em}
\begin{equation} \label{equation1}
    H = Att(Q,K,V)
\end{equation}
\vspace{-2em}

where $Att(\cdot)$ means the dot-product self-attention:
\vspace{-1em}
\begin{equation} \label{equation2}
    Att(Q,K,V) = softmax(Score_{Ori})V
\end{equation}
\vspace{-1.5em}
\begin{equation} \label{equation3}
    Score_{Ori} = \frac{QK^T}{\sqrt{d}}
\end{equation}
\vspace{-1em}

The original calculation regards all words as the same. To efficiently model and clarify trigger/argument field from the original ambiguous entanglement, we suppose that trigger $t_i$ or argument $a_i$ represents the central word of event mention $s$; further, we strengthen the importance of tokens in the field of 1-hop syntactic neighbors, which have the most critical syntactic information. Considering the distance contributions of different tokens to central words obey the normal distribution, we model event field via Gaussian prior in this paper.

Specifically, as shown in Figure~\ref{model}(a), the probabilistic bias $G \in \mathbb{R}^{n \times n}$, is defined to clarify and strengthen the original attention score of the event fields, i.e., $Score_{Ori}$ in Equation~\ref{equation2}:

\vspace{-1em}
\begin{equation}
    Att(Q,K,V) = softmax(Score_{Ori}+G)V
\end{equation}
\vspace{-1.75em}

The element $G_{i,j} \in (-\infty,0]$ measures the syntactic distance between context token $w_j$ in the 1-hop syntactic event field and the central trigger/argument word $w_i$, defined as:

\vspace{-1em}
\begin{equation}
    G_{i,j} = -\frac{(P_j-P_i)^2}{2{\sigma_i}^2}
\end{equation}
\vspace{-1em}

where $P_j$ and $P_i$ are the positions of neighbor and central words, $\sigma_i$ represents the standard deviation, defined as $\frac{D_i}{2}$, $D_i$ is the field size of the central word, i.e., the distance of 1-hop syntactic field to specific trigger/argument.



\noindent{\textbf{\underline{Event Field Recoupling.}}} Since the probabilistic bias was introduced independently for central word, it may be beneficial to consider them simultaneously when the same triggers/arguments appear multiple times in the mentions. As shown in Figure~\ref{model}, the trigger "attack" occurs twice in the mention, and they represent the same event type "war", the interaction between them could provide beneficial field-aware knowledge to each other.



Considering the words simultaneously appear in the interaction of several fields of the same triggers/arguments should be enhanced, we adopt Gaussian multiplication to recouple the clarified fields and obtain the latent distribution:

\vspace{-1em}
\begin{equation}
    G_{i,j} = -\frac{(P_j-P_i)^2}{2\sigma_i^2}-\frac{(P_j-\mu_{mul})^2}{2\sigma_{mul}^2}
\end{equation}
\begin{equation}
    \mu_{mul} = \sigma_{mul}^2\sum_{k=1}^{N}\frac{\mu_k}{{\sigma_k}^2},\quad\frac{1}{\sigma_{mul}^2} = \sum_{k=1}^{N}\frac{1}{{\sigma_k}^2}
\end{equation}
\vspace{-1em}

where $\mu_k$ and $\sigma_k$ denote the positions and syntactic field sizes of the same triggers/arguments, $\mu_{mul}$ and $\sigma_{mul}$ denote the latent probabilistic bias.


To combine several probabilistic distributions of the same triggers/arguments, we adopt GMM strategy to promote the recoupling of co-occurrence syntactic event fields, defined as:
\vspace{-1em}
\begin{equation}
    G_{i,j} = -\alpha\frac{(P_j-P_i)^2}{2{\sigma_i}^2}-\sum_{k=1}^{N}\beta_k\frac{(P_j-P_k)^2}{2{\sigma_k}^2}
\end{equation}
\vspace{-0.5em}

where $\alpha$ and $\beta_k$ denote the weights of original and other event fields of the same triggers/arguments. Considering exponential operation in $softmax$, we further define GMM as:
\vspace{-1em}
\begin{equation}
\begin{split}
    G_{i,j} = &\log(\frac{\alpha}{\sqrt{2\pi{\sigma_i}^2}}\exp{-\frac{(P_j-P_i)^2}{2{\sigma_i}^2}}\\&+\sum_{k=1}^{N}\frac{\beta_k}{\sqrt{2\pi{\sigma_k}^2}}\exp{-\frac{(P_j-P_k)^2}{2{\sigma_k}^2}})
\end{split}
\end{equation}
\vspace{-1em}

In this way, message passing among multiple distributions could be realized more efficiently.



\noindent{\textbf{\underline{Event Field Regularization Learning.}}} Considering most of the same triggers/arguments occurring multiple times in the same event mentions are identical, we hope the information learned from the identical event fields could be similar. Specifically, the corresponding distributions of predictions $P_k^{G_k}(y||x_i)$ should be more consistent, where $G_k$ means the $k_{th}$ probabilistic distribution of the same triggers/arguments.



Since the output predictions $P_j^{G_j}(y||x_i)$ and $P_k^{G_k}(y||x_i)$ of the same triggers/arguments specific fields enhanced knowledge are different, where $G_j$ and $G_k$ means the $j_{th}$ and $k_{th}$ probabilistic distribution of the same triggers/arguments, the predictions are different for the same input event mention $x_i$. To alleviate the inconsistency problem, we adopt a distribution divergence metric, i.e. WA Distance\footnote{where WA Distance could solve the asymmetry problem in Kullback-Leibler (KL) divergence, avoid the bidirectional calculation between two output distributions.}, to regularize the corresponding prediction distributions:

\vspace{-1.5em}
{\small
\begin{equation}
\begin{split}
    L_{WA}^i = \frac{\!\sum_{j=1}^{N-1}\!\sum_{k=j+1}^{N}\!WA(P_j^{G_j}(y||x_i),P_k^{G_k}(y||x_i))}{N(N-1)/2}
\end{split}
\end{equation}
}
\vspace{-1em}

With the negative log-likelihood learning objective $L_{NLL}^i$ of probabilistic enhanced predictions for given data $(x_i,y_i)$:

\vspace{-1em}
\begin{equation}
    L_{NLL}^i = -\frac{\sum_{j=1}^Nlog(P_j^{G_j}(y_i||x_i))}{N}
\end{equation}
\vspace{-1em}

Based on these training objectives, we define the final training objective as to minimize $L_{total}^i$ for the input event mention $x_i$:
\vspace{-1em}
\begin{equation}
\begin{split}
    L_{total}^i = L_{NLL}^i+L_{WA}^i
\end{split}
\end{equation}
\vspace{-2em}

In this way, our method further regularizes the model output distributions and enhances model's generalization. Besides, our method is compatible with the original BERT in model parameters and could be applied conveniently.
\section{Experiments}

\subsection{Setup}
\noindent{\textbf{\underline{Datasets.}}} We evaluate our methods on three generally-used EE datasets, including ACE 2005 English subset \cite{walker2006ace}, which contains 4,090 instances, 33 event types and 35 argument roles; the recently-constructed large-scale OntoEvent dataset \cite{deng2021ontoed}, which contains 60,546 instances and 100 event types; and FewEvent dataset \cite{deng2020meta}, which contains 70,852 instances and 100 event types. In the ACE 2005 dataset, we evaluate both Event Detection (ED) and Event Argument Classification (EAC) subtasks, while in OntoEvent and FewEvent, we only evaluate ED subtask due to their structures.




\noindent{\textbf{\underline{Baselines.}}} We adopt several official EE baselines, including: (1) CNN-based model DMCNN \cite{chen2015event}; (2) the model dependent on syntactic dependency knowledge, such as RNN-based model JRNN \cite{nguyen2016joint}, GCN-based model JMEE \cite{liu2018jointly}, graph-based models DYGIE++ \cite{wadden2019entity}, OneIE \cite{lin2020joint}, PathLM \cite{li2020connecting} and OntoED \cite{deng2021ontoed}, joint-based model Joint3EE \cite{nguyen2019one}; (3) GAN-based model GAIL \cite{zhang2019joint}; (4) some new forms of EE models, such as QA-based model BERT\_QA \cite{du2020event}, generation-based paradigm Text2Event \cite{lu2021text2event}; (5) BERT-based model DMBERT \cite{wang2019adversarial}.

\noindent{\textbf{\underline{General Settings.}}} We adopt the model structure of BERT, which is with 12 layers, 768 hidden sizes and 12 attention heads. AdamW optimizer is used with the learning rate of $1\times10^{-5}$. The dimension of token embedding is 768, while the maximum length of the input event mention is 128. The hyperparameters of $\alpha$ and $\beta$ are set to 0.5 and 0.5, respectively. We evaluate the performance of EE with Precision (P), Recall (R) and F1 Score (F1). We follow the evaluation protocol of previous EE models, event instances are split into training, validating and testing sets with the ratio of 0.8, 0.1 and 0.1, respectively. We run each method 5 times on all datasets and report the average performances to get stable results.

\subsection{Results and Analysis}
\noindent{\textbf{\underline{Overall Evaluation.}}} The evaluation results are shown in Tables~\ref{ACE} to~\ref{ONTO}. We can observe that:

(1) \emph{By modeling trigger/argument syntactic related fields, ProCE could efficiently clarify and recouple event fields from ambiguous entanglement.} ProCE achieves significant improvements compared with DMBERT on all datasets and outperforms all baselines, especially models based on syntactic dependency trees, such as JRNN, JMEE and Joint3EE.


(2) \emph{By combining different probabilistic enhancement strategies, ProCE could achieve different extents of improvements and utilize latent knowledge which benefits EE.} The general ablation study indicates that the recoupling interaction among probabilistic distributions could promote message passing, so the improvements of Gau Fusion, which combines Gau Mul and Gau GMM, are more significant than Gau. Gau Regularization could further improve the performance by strengthening the model's generalization.




\noindent{\textbf{\underline{Few-sample Evaluation.}}} Considering that auxiliary information will be more urgent in low-resource EE scenarios, we also study how to influence the ED performance of our method on the FewEvent dataset when trained with different ratios of randomly-sampled training data. We can see that:

\begin{table}
\centering
\resizebox{\linewidth}{!}{
\begin{tabular}{ccccccc}
\hline
\multirow{2}{*}{\textbf{Model}} & \multicolumn{3}{c}{\textbf{ED}} & \multicolumn{3}{c}{\textbf{EAC}}\\
\cline{2-7}
 & \textbf{P} & \textbf{R} & \textbf{F1} & \textbf{P} & \textbf{R} & \textbf{F1}\\
\hline
\multicolumn{1}{l}{DMCNN} & 75.60 & 63.60 & 69.10 & \textbf{62.20} & 46.90 & 53.50 \\
\multicolumn{1}{l}{JRNN} & 66.00 & 73.00 & 69.30 & 54.20 & 56.70 & 55.40 \\
\multicolumn{1}{l}{Joint3EE} & 68.00 & 71.80 & 69.80 & 52.10 & 52.10 & 52.10 \\
\multicolumn{1}{l}{DYGIE++} & - & - & 69.70 & - & - & 48.80 \\
\multicolumn{1}{l}{GAIL} & 74.80 & 69.40 & 72.00 & 61.60 & 45.70 & 52.40 \\
\multicolumn{1}{l}{OneIE} & - & - & 74.70 & - & - & 56.80 \\
\multicolumn{1}{l}{PathLM} & - & - & 73.40 & - & - & 56.60 \\
\multicolumn{1}{l}{BERT\_QA} & 71.12 & 73.70 & 72.39 & 56.77 & 50.24 & 53.31 \\
\multicolumn{1}{l}{Text2Event} & 69.60 & 74.40 & 71.90 & 52.50 & 55.20 & 53.80 \\
\hline
\multicolumn{1}{l}{DMBERT} & 71.60 & 72.30 & 70.87 & 53.14 & 54.24 & 52.76 \\
\multicolumn{1}{l}{+Gau} & 73.43 & 74.23 & 72.98 & 53.82 & 54.45 & 53.22 \\
\multicolumn{1}{l}{+Fusion} & 74.22 & 76.24 & 74.70 & 56.35 & 55.12 & 54.27 \\
\multicolumn{1}{l}{+Regularization} & \textbf{77.16} & \textbf{77.96} & \textbf{76.80} & 57.43 & \textbf{58.37} & \textbf{57.03} \\
\hline
\end{tabular}}
\caption{\label{ACE}
Evaluation of EE with various models on ACE2005.
}
\end{table}

\begin{table}
\centering
\begin{tabular}{ccccc}
\hline
\multirow{2}{*}{\textbf{Model}} & \multicolumn{3}{c}{\textbf{ED}} \\
\cline{2-4}
 & \textbf{P} & \textbf{R} & \textbf{F1} \\
\hline
\multicolumn{1}{l}{DMCNN} & 62.51 & 62.35 & 63.72 \\
\multicolumn{1}{l}{JRNN} & 63.73 & 63.54 & 66.95 \\
\multicolumn{1}{l}{JMEE} & 52.02 & 53.80 & 68.07 \\
\multicolumn{1}{l}{AD-DMBERT} & 67.35 & 73.46 & 71.89 \\
\multicolumn{1}{l}{OneIE} & 71.94 & 68.52 & 71.77 \\
\multicolumn{1}{l}{PathLM} & 73.51 & 68.74 & 72.83 \\
\multicolumn{1}{l}{OntoED} & 75.46 & 70.38 & 74.92 \\
\hline
\multicolumn{1}{l}{DMBERT} & 80.00 & 78.30 & 78.40 \\
\multicolumn{1}{l}{+Gau} & 80.45 & 79.80 & 79.70 \\
\multicolumn{1}{l}{+Fusion} & 81.18 & 80.32 & 80.30 \\
\multicolumn{1}{l}{+Regularization} & \textbf{82.30} & \textbf{80.57} & \textbf{80.90} \\
\hline
\end{tabular}
\caption{\label{ONTO}
Evaluation of ED on OntoEvent.
}
\vspace{-1em}
\end{table}

(1) \emph{ProCE is especially beneficial for extremely low-resource EE scenarios.} As shown in Table~\ref{FewSample}, the improvements of our methods compared to the baseline DMBERT are generally more significant when less training data is available. Specifically, in extremely low-resource EE scenarios (training with less than 20\% data), Gau Fusion obtains 37.23\% F1 with 10\% training data, while Gau Regularization obtains 47.13\% F1, in comparison to 34.93\% in DMBERT.

(2) \emph{ProCE could achieve better performance with less data constantly.} As shown in Table~\ref{FewSample}, ProCE obtains more advanced performances with less training data continuously. Especially, DMBERT requires 100\% training data to almost achieve the best performance, while Gau Fusion only needs 80\%. Gau Fusion could even obtain 81.65\% F1 with overall data, while Gau Regularization obtains 82.43\%, 1.91\% higher than 80.52\% in DMBERT.


\label{sec:bibtex}



\begin{table}
\centering
\resizebox{\linewidth}{!}{
\begin{tabular}{cccccccc}
\hline
\multirow{2}{*}{\textbf{Model}} & \multicolumn{7}{c}{\textbf{ED}} \\
\cline{2-8}
 & \textbf{5\%} & \textbf{10\%} & \textbf{20\%} & \textbf{40\%} & \textbf{60\%} & \textbf{80\%} & \textbf{100\%} \\
\hline
\multicolumn{1}{l}{DMBERT} & 24.37 & 34.93 & 56.27 & 72.47 & 78.17 & 79.10 & 80.52 \\
\multicolumn{1}{l}{+Gau} & 24.93 & 35.80 & 56.48 & 72.77 & 78.30 & 79.75 & 81.30 \\
\multicolumn{1}{l}{+Fusion} & 25.97 & 37.23 & 57.25 & 73.67 & 79.03 & 80.23 & 81.65 \\
\multicolumn{1}{l}{+Regularization} & \textbf{32.03} & \textbf{47.13} & \textbf{62.47} & \textbf{75.53} & \textbf{80.13} & \textbf{81.17} & \textbf{82.43} \\
\hline
\end{tabular}}
\caption{\label{FewSample}
Few-sample evaluation with F1-score $(\%)$ performance of ED with various models on FewEvent.
}
\end{table}

\begin{table}
\centering
\resizebox{\linewidth}{!}{
\begin{tabular}{cccccccccc}
\hline
\multirow{2}{*}{\textbf{Model}} & \multicolumn{3}{c}{\textbf{ACE ED}} & \multicolumn{3}{c}{\textbf{ONTO ED}} & \multicolumn{3}{c}{\textbf{FEW ED}}\\
\cline{2-10}
 & \textbf{P} & \textbf{R} & \textbf{F1} & \textbf{P} & \textbf{R} & \textbf{F1} & \textbf{P} & \textbf{R} & \textbf{F1}\\
\hline
\multicolumn{1}{l}{DMBERT} & 71.60 & 72.30 & 70.87 & 80.00 & 78.30 & 78.40 & 81.75 & 81.83 & 80.52 \\
\multicolumn{1}{l}{+Gau} & 73.43 & 74.23 & 72.98 & 80.45 & 79.80 & 79.70 & 82.27 & 82.58 & 81.30 \\
\multicolumn{1}{l}{+Mul} & 73.39 & 75.24 & 73.56 & 81.16 & 80.29 & 80.13 & 82.48 & 82.85 & 81.55 \\
\multicolumn{1}{l}{+GMM} & 73.18 & 74.78 & 73.28 & 81.00 & 79.98 & 79.85 & 82.78 & 82.62 & 81.46 \\
\multicolumn{1}{l}{+Fusion} & 74.22 & 76.24 & 74.70 & 81.18 & 80.32 & 80.30 & 82.53 & 83.00 & 81.65 \\
\multicolumn{1}{l}{+WA} & 76.52 & \textbf{78.00} & 76.30 & 81.90 & 80.10 & 80.23 & \textbf{83.57} & 82.83 & 81.83 \\
\multicolumn{1}{l}{+WA \& logits} & \textbf{77.16} & 77.96 & \textbf{76.80} & \textbf{82.30} & \textbf{80.57} & \textbf{80.90} & 83.17 & \textbf{83.80} & \textbf{82.43} \\
\hline
\end{tabular}}
\caption{\label{ablation}
Ablation study of EE on three datasets.
}
\vspace{-1em}
\end{table}

\noindent{\textbf{\underline{Detailed Ablation Analysis.}}} To evaluate the effect of different probabilistic-based enhancement methods, we study the performances of the corresponding modules. The ablation results are shown in Table~\ref{ablation}. We can observe that:

(1) \emph{Event Field Recoupling strategy could guide the generation of more syntactic knowledge.} The F1 of Gau is 1.16\% higher than DMBERT on the average of all datasets. The Gau Mul slightly surpasses the Gau GMM since Gau Mul generates more unseen information. The fusion of Gau Mul and Gau GMM indicates that promoting message passing among several identical event fields could further benefit the model. The F1 of Gau Fusion is 0.93\% higher than Gau on average.

(2) \emph{Event Field Regularization strategy is essential for model's generalization.} The comparison between Gau WA and Gau WA \& logits shows the importance of adjusted probabilistic logit guidance. Since Gau WA only relies on the regularization of probabilistic prediction distributions, the improvements are implicit; Gau WA even results in a 0.07\% performance drop by comparison with Gau Fusion on the OntoEvent dataset. Gau WA \& logits utilizes all probabilistic enhanced logits to further improve regularization. The F1 of Gau WA \& logits is 0.49\% higher than Gau WA on the average of all datasets, 3.65\% higher than DMBERT on average.


\section{Conclusion and Future Work}
\label{sec:bibtex}

In this paper, we propose ProCE, a probabilistic recoupling enhanced EE framework, to model the syntactic-related event field knowledge for the first time. To explore the latent entanglement among probabilistic distributions, we construct event field recoupling and regularization methods. Experiments on several datasets demonstrate that our model achieves significant improvements above the previous state-of-the-art models, especially those based on PLMs and syntactic knowledge. Further experiments in few-sample scenarios indicate that our model benefits low-resource EE tasks.

We will expand our methods to other keyword-based IE tasks, such as relation extraction, to improve the generalization ability of our model for future work.






\vfill\pagebreak

\bibliographystyle{IEEEbib}
\bibliography{refs}

\end{document}